\documentclass[letterpaper, 10 pt, conference]{ieeeconf}  

\IEEEoverridecommandlockouts                              

\usepackage{graphicx}
\usepackage{amsmath}
\usepackage{amssymb}
\usepackage{setspace}
\usepackage{algpseudocode}
\usepackage{algorithm}
\let\Algorithm\algorithm
\renewcommand\algorithm[1][]{\Algorithm[#1]\setstretch{1.3}}

\usepackage{color}  

\title{\LARGE \bf
Particle-based Pedestrian Path Prediction using LSTM-MDL Models
}

\author{Ronny Hug$^{*}$, Stefan Becker$^{*}$, Wolfgang H\"ubner$^{*}$ and Michael Arens$^{*}$
\thanks{*Fraunhofer Institute of Optronics, System Technologies and Image Exploitation IOSB,
Gutleuthausstr. 1, 76275 Ettlingen, Germany. E-mails:
{\tt\small \{ronny.hug, stefan.becker, wolfgang.huebner, michael.arens\}@iosb.fraunhofer.de}}
}

\begin{document}

\maketitle
\thispagestyle{empty}
\pagestyle{empty}

\begin{abstract}

Recurrent neural networks are able to learn complex long-term relationships from sequential data and output a probability density function over the state space. 
Therefore, recurrent models are a natural choice to address path prediction tasks, where a trained model is used to generate future expectations from past observations.   
When applied to security applications, like predicting pedestrian paths for risk assessment, a point-wise greedy evaluation of the output pdf is not feasible, since the environment often allows multiple choices. 
Therefore, a robust risk assessment has to take all options into account, even if they are overall not very likely.   

Towards this end, a combination of particle filtering strategies and a LSTM-MDL model is proposed to address a multi-modal path prediction task. 
The capabilities and viability of the proposed approach are evaluated on several synthetic test conditions, yielding the counter-intuitive result that the simplest approach performs best.
Further, the feasibility of the proposed approach is illustrated on several real world scenes.

\end{abstract}

\section{Introduction}

\begin{figure}[t]
	\begin{center}		
		\includegraphics[width=0.375\textwidth]{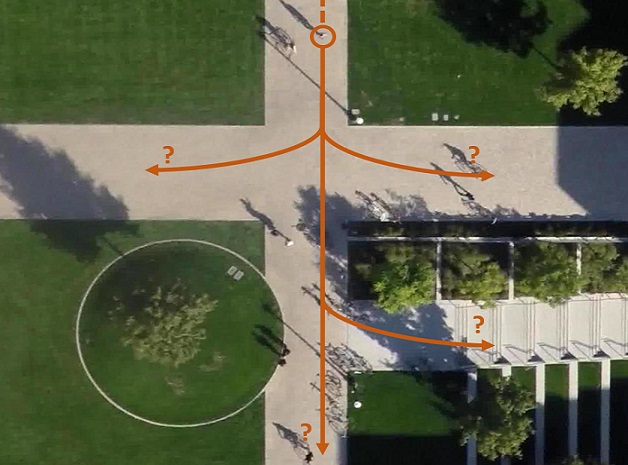}
	\end{center}
	\vspace{-0.2cm}
	\caption{Multi-modal path prediction: Where might the person in the orange circle be in $N$ time steps?}
	\label{fig:problem}
	\vspace{-0.28cm}
\end{figure}

A common task in the context of intelligent vehicles is risk assessment, where a risk score is calculated from interpreting a set of possible future paths as generated from a path prediction method. 
The risk score can e.g. encode the chance of a collision between a pedestrian and a vehicle for some point in the future, where strong prediction capabilities are required in order to enhance the time horizon where a useful prediction is possible. 
Basing decisions on anticipated pedestrian behavior yields the advantage of earlier decisions, especially in time critical situations.
Further, the generation of multi-modal predictions is an essential capability, since environments, like the one depicted in Fig. \ref{fig:problem} commonly offer multiple choices, which are not equal likely but in general none of the options is totally unlikely.

In contrast to prediction approaches, taking only the motion state of an object into account, path prediction in dynamic environments can strongly benefit from a complex transition or motion model, capable of capturing motion decisions of pedestrians, including:
\begin{enumerate}
	\item Long-term dependencies, where decisions are influenced from past motion in a non-linear fashion
	\item The inclusion of the scene geometry, e.g. implicitly included in the statistical model (\cite{huang2016deep})
	\item Local interaction with dynamic objects in the scene (\cite{bartoli2017context}\cite{alahi2016social})
	\item Decision making on different time scales, separating local and global path planning
	\item Goal oriented motion, mainly summarized as the intuition of a pedestrian
\end{enumerate}

While the intuition of observed pedestrians is hardly measurable, the other components can either be observed or learned from data when using a statistical transition model.
A class of flexible statistical models that can be used on varying time scales are recurrent neural networks.
For multi-modal path prediction, these models can be built, such that a probability density function over the state space is generated.

Approaches dealing with path prediction can roughly be categorized by the targeted type of path prediction, which is goal-directed (\emph{what is the most likely path towards a given destination}) or undirected path prediction (\emph{where will the observed pedestrian be in $N$ time steps}).
Since the goal location is usually unknown, this paper is mainly concerned with undirected path prediction.
While recent undirected path prediction approaches are only concerned with unimodal or example-based multi-hypothesis prediction, multi-modal path prediction is only subject to goal-directed prediction approaches.
Towards this end, the goal of this paper is to introduce an undirected, multi-modal path prediction approach based on techniques known from particle filtering.

In the following, a brief summary of related work and the recurrent model used for path prediction are provided (section \ref{s:background}).
Next, the proposed path prediction approach is presented in section \ref{s:path_pred}.
The quality of the generated probability distribution is evaluated on different synthetic and real-world conditions (section \ref{sec:exp}).
Section \ref{s:conclusion} provides some concluding remarks.

\section{Background}
\label{s:background}

The path prediction problem has been addressed with a variety of approaches based on dynamic models, like recurrent neural networks (e.g. \cite{bartoli2017context}\cite{alahi2016social}) or Markov processes (e.g. \cite{kitani2012activity}), or on just a single observation (e.g. \cite{huang2016deep}).
In the case of goal-directed prediction, a policy optimization is performed after processing observations.
For undirected predictions, a state-transition model is used to project given information into the future.
Further, path prediction can be approached on a local and a global scale.
On a global scale, a path taken by a pedestrian is mainly influenced by the intended destination (indicated by observed past positions) and the scene geometry (e.g. \cite{bartoli2017context}\cite{kitani2012activity}).
On a local scale, dynamic obstacles come into play and alter the intended path (e.g. \cite{alahi2016social}).
There are also hybrid approaches, targeting both scales of path prediction (e.g. \cite{lee2017desire}).
In the context of risk assessment, global-scale path prediction should be considered for long-term decisions and complemented by local-scale path predictions for short-term decisions, respectively.
The latter is especially relevant, as pedestrian interactions may lead to risky situations.

Although being an inference problem, path prediction can also be construed as a sequence generation problem by applying recurrent neural networks (RNNs).
One of the main advantages in using RNNs is their capability of processing and generating sequences of variable lengths.
While simple RNNs struggle in capturing long-term dependencies, gated memory blocks, like the long short-term memory (LSTM, \cite{hochreiter1997long}) or gated recurrent unit (GRU, \cite{cho2014learning}), can be used.
Further, when combining the RNN with a mixture density layer (MDL, \cite{bishop2006pattern}), the model can be trained to output a continuous probability distribution.
Here, undirected, multi-modal path prediction is based on a LSTM model that is combined with a MDL, due to the flexibility of recurrent models and the probabilistic output.

\subsection{The LSTM-MDL model}
\label{s:model}
The LSTM-MDL model \cite{graves2013generating} is a recurrent neural network (using LSTM cells\footnote{It could as well be built using Gated Recurrent Units.}) that is used to parameterize a mixture density output layer (MDL).
The model is trained by minimizing the negative log-likelihood loss.
Given an input sequence $\mathcal{X}^T = \{\mathbf{x}^1, \mathbf{x}^2, ..., \mathbf{x}^T\}$ of $T$ consecutive pedestrian positions along a trajectory, the model generates a, at least $K$-modal, prediction for the next position $\mathbf{x}^{t+1}$ or position offset $\delta^t$ at each time step.
This prediction of a model $\mathcal{M}$ at time $t$ is a $K$-component Gaussian mixture model (GMM) defined by parameters $\pi_k$, $\mu_k$ and $\Sigma_k$ (weight, mean and covariance of the k'th Gaussian):
\begin{align}
\label{eq:gmm}
\begin{split}
\mathcal{M}(\mathbf{x}^t) &= (\pi^t, \mu^t, \Sigma^t) \\
&\Rightarrow p(\mathbf{\delta}|\mathbf{x}^t, \pi^t, \mu^t, \Sigma^t) \\
&\overset{\wedge}{=} p(\mathbf{\delta}^t|\mathbf{x}^t) = \sum_{k=1}^{K} \pi_k \mathcal{N}(x|\mu_k,\Sigma_k),
\end{split}
\end{align}

\subsection{Path Prediction using the LSTM-MDL model}
Several approaches base undirected path prediction on this architecture, but are limited to unimodal predictions.
For example in \cite{bartoli2017context}\cite{alahi2016social}, the LSTM-MDL model serves as an egocentric pedestrian motion model and interaction was included in order to improve prediction results.
For the purpose of this paper, however, a simpler model is used to reduce necessary amounts of training data and further to reduce side-effects while evaluating the proposed prediction approach.
Therefore, the model presented in \cite{hug2017reliability}, which does not regard any context (e.g. neighboring pedestrians or the scenery), is used.

Even though the LSTM-MDL model outputs a probability distribution, the model itself is inherently deterministic.
Because of this, the model has to be embedded in an inference scheme that explores the probability distributions generated by the model.
Due to the models property of processing each time step in succession, inference needs to follow an iterative approach and prompt new positional distribution from the model in each time step.
Further, this structure leads to a series of conditional distributions, which is why global samplers, like Gibbs sampling \cite{bishop2006pattern}, cannot be applied.

The approach commonly used to generate predictions using a trained LSTM-MDL model $\mathcal{M}$ that generates a GMM $p(\mathbf{\delta}^t|\mathbf{x}^t)$ describing the offset distribution to the next position (see (\ref{eq:gmm})) is as follows (for each time step): 
\begin{enumerate}
	\item Generate $p(\mathbf{\delta}^t|\mathbf{x}^t)$ by passing the current position $\mathbf{x}^t$ through $\mathcal{M}$
	\item Transform offset distribution into a positional distribution $p(\mathbf{x}^{t+1}|\mathbf{x}^t)$ by adding $\mathbf{x}^t$ to every $\mu^t_k \in \mu^t$
	\item Sample a position $\hat{\mathbf{x}}^{t+1}$ from $p(\mathbf{x}^{t+1}|\mathbf{x}^t)$
	\item Use $\hat{\mathbf{x}}^{t+1}$ in step 1 to generate the next prediction.
\end{enumerate}
Most commonly, $\hat{\mathbf{x}}^{t+1}$ maximizes $p(\cdot|\mathbf{x}^t)$, thus generating a maximum likelihood solution, disregarding less likely paths.

To achieve a multi-modal prediction, the pdf output of the model has to be fed back into it (as opposed to a single position) in order to move forward in time.
Further, the LSTM-MDL model expects single positions as input, thus an efficient sample-based approximation of the probability distribution has to be used.
This, in turn, may lead to exponential growth in the number of samples and the number of LSTM cell states.
In total this leads to complex conditional distributions which are usually intractable in a straightforward way.

\subsection{Particle-based inference on recurrent models}

To cope with these problems, a particle-based prediction approach is proposed.
Towards this end, common particle filtering algorithms are adapted for usage on top of a LSTM-MDL model.
Thereby, different well-established techniques in the area of particle filtering were incorporated and evaluated in terms of applicability and usefulness.
Although using a particle-based approach for inference on RNNs is no novelty when looking at state-space models \cite{gu2015neural}\cite{zheng2017state} concerned with modeling latent space, there is a fundamental difference between these models and the LSTM-MDL model.
While the aforementioned state-space models are themselves stochastic in their latent space and monte carlo simulation is applied in the training phase, the LSTM-MDL model itself is deterministic and turns into a probabilistic model by embedding it into a monte carlo simulation when performing prediction. 

\subsection{Problem description}
\label{s:problem}

In the following, the goal is to generate a probabilistic prediction for the next $N$ positions $\{p(\mathbf{x}^{T+1}), p(\mathbf{x}^{T+2}), ..., p(\mathbf{x}^{T+N})\}$, given an observation $\mathcal{X}^T = \{\mathbf{x}^1, \mathbf{x}^2, ..., \mathbf{x}^T\}$ of $T$ consecutive pedestrian positions along a trajectory.
In order to explore all likely paths the model is aware of, the GMMs $p(\mathbf{x}^{t+1}|\mathbf{x}^t)$ have to be fed back into $\mathcal{M}$ to progress in time at each time step.
As the model requires positions as input, $p(\mathbf{x}^{t+1}|\mathbf{x}^t)$ can be represented by a sufficiently large number of samples, which are then passed through $\mathcal{M}$.
The distributions $p_s(\mathbf{x}^{t+2}|\mathbf{x}^{t+1})$ generated by each sample $s$ can then be used to draw the next set of samples and so on.
Following this na\"ive approach, the number of samples increases exponentially with each time step, making this exploration prohibitive in terms of computation time and memory consumption.
Additionally, each sample is associated with a separate LSTM cell state, thus the number of states also grows exponentially.

\section{Particle-based path prediction}
\label{s:path_pred}

A more sophisticated approach to approximate a series of distributions using a fixed number of weighted samples is taken in particle filtering \cite{arulampalam2002tutorial}.
Here, the samples (particles) are updated in each time step considering a simple motion model and frequent measurements.
In the process of predicting the positional distribution at time $t+n$, the motion model could be applied recursively.

On a considerable high level, the particle filtering scheme can be adapted for use in multi-modal path prediction using a LSTM-MDL model as follows.
A set $\mathcal{P}^t = \{\mathbf{p}^t_1, \mathbf{p}^t_2, ..., \mathbf{p}^t_M\}$ of $M$ particles with corresponding weights $\Omega^t = \{\omega^t_1, \omega^t_2, ..., \omega^t_M\}$ is used to describe $p(\mathbf{x}_{t}|\mathbf{x}_{t-1})$ at each time step $t$.
The LSTM-MDL model serves as the motion model to propagate the set of particles forward in time.
When passing $\mathcal{P}^t$ through $\mathcal{M}$, a set of $M$ GMMs (see (\ref{eq:gmm}))
\begin{align}
\mathcal{M}(\mathcal{P}^t) \Rightarrow \mathcal{G}^{t+1} = \{p_m(\cdot|\mathbf{p}^t_m) \mid \mathbf{p}^t_m \in \mathcal{P}^t\},
\end{align}
one per particle, is generated.
To allow a fixed number of particles, all distributions in $\mathcal{G}^{t+1}$ need to be aggregated into a single $K \cdot M$ - component GMM
\begin{align}
\label{eq:comb_gmm}
p(\mathbf{x}^{t+1}|\mathcal{G}^{t+1}) = \sum_{m=1}^{M} \omega^t_m \cdot p_m.
\end{align} 
Here, it is vital that $\mathcal{M}$ is an accurate model of various paths, as there are no incoming measurements to correct intermediate predictions.
Further, without new measurements, it has to be clarified how to determine the particle weights $\Omega$.
This step is of great importance, as the particle weights affect the aggregation of $\mathcal{G}^{t+1}$.
Additionally, due to the fact that the LSTM-MDL model cannot be used to update given particles, the usual particle update step known from the particle filter has to be replaced by completely re-sampling all particles in each time step given $p(\mathbf{x}^{t+1}|\mathcal{G}^{t+1})$.

\begin{algorithm}[h]
	\begin{algorithmic}[1]
		\Require $\mathcal{X}^T,~N,~\mathcal{M}$ \Comment{observation $\mathcal{X}^t$, model $\mathcal{M}$}
		\State $p(\cdot|\mathbf{x}^T) \gets precondition(\mathcal{M}, \mathcal{X}^t)$ \Comment{$K$ components}
		\State $\mathcal{P}^1 \gets draw\_samples(p(\cdot|\mathbf{x}^T))$ 
		\State $\Omega^1 \gets weight\_particles(\mathcal{P}^1, p(\cdot|\mathbf{x}^T))$
		\For{$s \gets [2..N]$} \Comment{propagate particles}
		\State $\mathcal{G}^s \gets \mathcal{M}(\mathcal{P}^{s-1})$ \Comment{$\{p_m(\cdot|\mathbf{p}^{s-1}_m) \mid \mathbf{p}^{s-1}_m \in \mathcal{P}^{s-1}\}$}
		\State $p(\cdot|\mathcal{G}^{s}) \gets \sum_{m=1}^{M} \omega^{s-1}_m \cdot p_m$ \Comment{$M \cdot K$ components}
		\State $\mathcal{P}^{s} \gets draw\_samples(p(\cdot|\mathcal{G}^{s}))$
		\State $\Omega^s \gets weight\_particles(\mathcal{P}^{s}, p(\cdot|\mathcal{G}^{s}))$
		\EndFor
		\State \textbf{return} $p(\cdot|\mathcal{G}^{s}),~\mathcal{P}^s$
	\end{algorithmic}
	\caption{Particle Propagation}\label{alg:particle_prop}
\end{algorithm}
The proposed prediction method is summarized in algorithm \ref{alg:particle_prop}.
The algorithm starts given an observation sequence $\mathcal{X}^T$, the number $N$ of time steps to predict and the trained LSTM-MDL model $\mathcal{M}$.
In the initialization stage (lines 1 to 3), the model is conditioned on the observations by consecutively passing each position $\mathbf{x}^t \in \mathcal{X}^T$ through $\mathcal{M}$, while updating the LSTM cell states.
The transformed model output at time step $T$ $p(\cdot|\mathbf{x}^T)$ (cf. (\ref{eq:gmm})) describes the first predicted positional distribution.
From here, the steps described above are performed in a loop to produce the sequence of conditional distributions $\{p(\mathbf{x}^{T+1}|\cdot), p(\mathbf{x}^{T+2}|\cdot), ..., p(\mathbf{x}^{T+N}|\cdot)\}$ (see section \ref{s:problem}).
It has to be noted, that in step 5, where the offset distribution is transformed, the particle that generated the corresponding distribution has to be used.
I.e. $p_m(\mathbf{x}^{t+1}|\mathbf{p}^t_m)$ is obtained from $p_m(\mathbf{\delta}^{t}|\mathbf{p}^t_m)$ by adding $\mathbf{p}^t_m$ to the mean vectors.
Further, new particles inherit the LSTM state from their ancestors.

The following sections break down the algorithms main parts: how to sample particles from a given GMM (section \ref{sss:sampling}) and how to weight particles (section \ref{sss:weighting}).
The presented strategies are intended to address the particle set degeneration problem common to particle filtering, where all particles collapse into a dense region, thus preventing the exploration of different paths.
As it is unclear which strategies will be adequate for the proposed approach, different techniques are investigated.

\subsection{Sampling strategies}
\label{sss:sampling}

The common approach for sampling from a GMM $p(x) = \sum_{k=1}^{K \cdot M} \pi_k \mathcal{N}(x|\mu_k,\Sigma_k)$, is to select one of the $K \cdot M$ Gaussian components and then draw a sample from that Gaussian distribution.
The strategies described in this section are concerned with the selection (i.e. the index $k$) of the Gaussian component to sample from.
Usually, $k$ is selected by performing a multinomial sampling given the component weights.
In particle filtering, there are other commonly used sampling approaches to tackle particle set degeneration.
Here, a degenerating particle set results in less likely paths vanishing from the prediction.
Prominent examples are systematic \cite{kitagawa1996monte} and stratified \cite{fishman2013monte} sampling.
Due to their similar impact on the sampling results \cite{hol2006resampling}, only stratified sampling is described and evaluated in the remainder of this paper.

\paragraph{Multinomial sampling} 
When using a multinomial sampling approach, the Gaussian component to sample from is determined by drawing $k$ from a multinomial distribution, where each $k \in [1..(K \cdot M)]$ is one possible outcome and each $k$ is drawn with probability $\pi_k$.
Having chosen $k$, a sample is drawn from $\mathcal{N}(\mu_k,\Sigma_k)$.
This is repeated $M$ times.

\paragraph{Stratified sampling}
While multinomial sampling may lead to only few of the $K \cdot M$ components being selected when drawing multiple samples in succession (due to small weights on some components), stratified sampling aims at drawing $k$ more uniformly from the weights $\pi_k$.
In theory, this leads to less probable components being chosen more frequently.
This is achieved by first calculating the cumulative sum of the weights $\pi_k$.
Then, the sum is subdivided into $C$ equally sized bins.
Here, $C=M$ is the number of samples to be drawn.
Each of these sections refers to one or more values for $k$.
Lastly, for each section a component $k$ is chosen, such that $k = l^k + u$, where $u \sim \mathcal{U}(\left[0, 1\right])$ and $l^k$ and $r^k$ are the values of the left and right border of the $k$'th bin.
As before, samples are drawn from the corresponding Gaussian components.

\subsection{Particle weighting}
\label{sss:weighting}

According to common particle filtering, a weight needs to be assigned to each particle.
Here, these weights determine how $\mathcal{G}^t$ is combined into a single GMM $p(\cdot|\mathcal{G}^{t})$ (cf. (\ref{eq:comb_gmm})).
When assigning no weight to the particles, all GMMs $p_m(\cdot|\mathbf{p}^{t-1}_m) \in \mathcal{G}^t$ will be combined equally-weighted, thus
\begin{align}
\omega^t_m = \frac{1}{M},~\forall m \in [1..M].
\end{align} 
As a consequence, there are more particles in regions of high probability, leading to an indirect weighting.
By assigning different weights to the particles, the probability density of different regions in $p(\cdot|\mathcal{G}^{t})$ can be increased or decreased to some degree.
Thus, more emphasis can be put on the largest mode in the distribution, eventually leveling out secondary peaks.
Conversely weak regions can be pushed in order to prevent a collapse onto a single peak.
Pushing weak regions might help in exploring less likely paths.

In the following, several weighting strategies are presented.
The first strategy, \emph{density value weighting}, yields a straightforward way to assign a weight to each particle by using $p(\cdot|\mathcal{G}^{t})$ (cf. (\ref{eq:comb_gmm})).
The strategies \emph{temperature weighting} and \emph{interpolation weighting} can be used to shift the weight distribution achieved by \emph{density value weighting}.
Depending on the parameters, these strategies are intended to force exploration of less likely paths.

\paragraph{Density value weighting}
A straightforward way of weighting the particles is to apply the pdf $p(\cdot|\mathcal{G}^{t})$ (cf. (\ref{eq:comb_gmm})) that was used to produce the particle set, by passing each particle $\mathbf{p}^t_m$ through it
\begin{align}
\omega^t_m = p(\mathbf{p}^t_m|\mathcal{G}^{t}).
\end{align} 
Following that, the set of weights $\Omega^t$ has to be normalized, to conform $\sum^M_{m=1} \omega^m = 1$.

\paragraph{Temperature weighting}
First, the particle weights are determined using density value weighting.
After that, the weight distribution is shifted by applying the transformation
\begin{align}
\begin{split}
	\omega^t_{m, temp} = \frac{e^{\frac{ln \omega^t_m}{\tau}}}{\sum^M_{i=1} e^{\frac{ln \omega^t_i}{\tau}}} = \frac{(\omega^t_m)^{\frac{1}{\tau}}}{\sum^M_{i=1} (\omega^t_i)^{\frac{1}{\tau}}}
\end{split}
\end{align}
to each weight $\omega^t_m$ for $m \in [1..M]$.
The parameter $\tau$ is referred to as the temperature.
For $\tau \rightarrow 0$, the new weights $\Omega^t_{temp}$ are shifted towards the strongest component, while for $\tau \rightarrow \infty$, $\Omega^t_{temp}$ approaches an uniform distribution.
This transformation is a slight variation of the \emph{temperature softmax}, sometimes used in reinforcement learning \cite{sutton1998reinforcement}.
It deviates from the original formula in that the natural logarithm is applied to each $\omega^t_m$.
This variation adds the identity transformation $\Omega^t_{temp} = \Omega^t$ for $\tau = 1$.

\paragraph{Interpolation weighting}
Temperature weighting allows to put more emphasis on the largest weights or to approach an uniform distribution.
Interpolation weighting allows to put more emphasis on smaller weights as well.
This is achieved by linearly interpolating each weight $\omega^t_m$ for $m \in [1..M]$ using an interpolation factor $\kappa$:
\begin{align}
\omega^t_{m, interp} = (1 - \kappa) * \omega^t_m + \kappa * (1 - \omega^t_m).
\end{align}
Again, the set of weights has to be re-normalized.
Using this approach, $\kappa = 0$ does not change the weights, $\kappa = 0.5$ results in an uniform distribution and $\kappa \in (0.5, 1]$ puts more emphasis on smaller weights.
For $\kappa \in (0, 0.5)$, this approach yields results similar to temperature weighting when $\tau \rightarrow \infty$.

The motivation to strengthen less probable components is to help paths that may be highly probable in later time steps to \emph{survive} early stages of the prediction.
Still, from the initial set of particles, there are more particles from high probability regions than there are from low probability regions, thus the most relevant paths are not lost, even with $\kappa = 1$.
These weights only affect how single GMMs are combined and not individual component weights, thus there is no risk that components with weights close to $0$ are enhanced.

\section{Experiments}
\label{sec:exp}

This section shows a quantitative (\ref{ss:quant}) and a qualitative (\ref{ss:quali}) evaluation for the proposed approach.
Both evaluations are concerned with verifying the overall viability of the approach in different situations given different sets of parameters.
Assuming that the LSTM-MDL model is capable of modeling all relevant paths in the training data \cite{hug2017reliability}, the quantitative evaluation is concerned with the difference between the probability distribution generated by the predictor using different configurations and the true distribution as provided by the training data.
A set of synthetic test conditions is used in order to reduce the number of factors that might cause the predictor to bias into a certain path or direction and to allow for a more systematic evaluation.
In the qualitative evaluation part, the performance of the best (in terms of the quantitative evaluation) predictor is shown on real-world scenes.

\begin{figure*}[t]
	\centering  
	\begin{tabular}{c@{\hspace{1pt}}c@{\hspace{2pt}}c@{\hspace{1pt}}c}
		\includegraphics[width=0.24\textwidth]{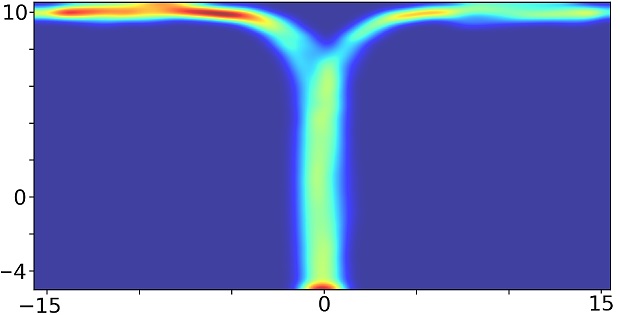} & \includegraphics[width=0.24\textwidth]{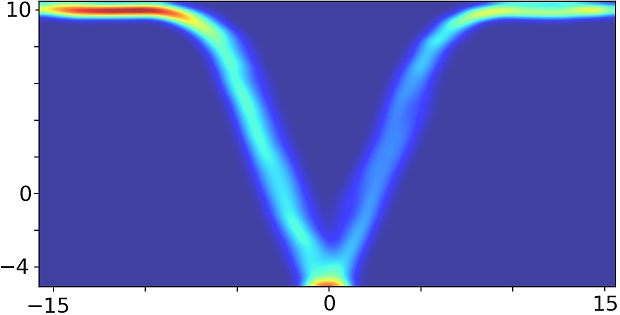} & \includegraphics[width=0.24\textwidth]{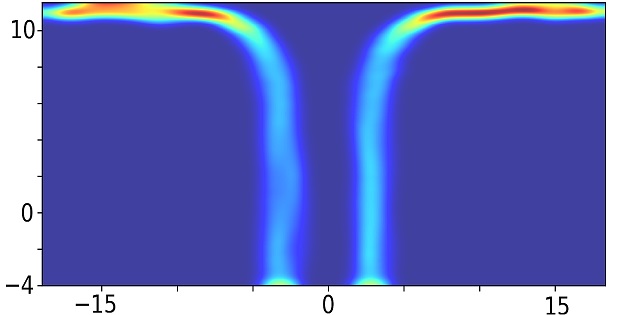} & \includegraphics[width=0.24\textwidth]{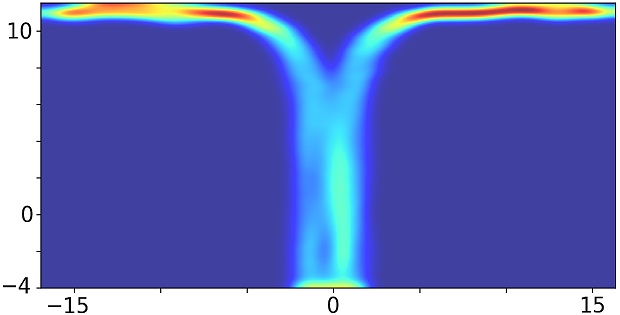} \\
		\small{1: \emph{tmaze-heavy-left}} & \small{2: \emph{tmaze-dirbias}} & \small{3: \emph{tmaze-posbias-gap}} & \small{4: \emph{tmaze-posbias-nogap}}
	\end{tabular}
	\caption{Density plot for trajectories in synthetic test conditions (all trajectories start at the bottom, red = high density).} 
	\label{fig:synth_data}
	\vspace{-0.2cm}
\end{figure*}
\subsection{Implementation details}
The LSTM-MDL model and the prediction algorithm have been implemented using tensorflow.
The implementation strongly utilizes vectorized calculations performed on the GPU, which come at the cost of higher memory consumption, thus limiting the maximum number of particles for the prediction.
On the test system (Intel Core i7-5930K, 32GB RAM, Nvidia Titan X), the number of particles is capped at 5000.
For the quantitative evaluation, each prediction has been performed 10 times and particles have been aggregated to a total number of 50000 particles per prediction.

\subsection{Quantitative results}
\label{ss:quant}

For the quantitative evaluation, several variations of noisy trajectories along a t-shaped junction have been generated synthetically.
A t-shape can be considered as a prototypical decision scenario that is resembled in various forms in the real world (e.g. see Fig. \ref{fig:qual_res}).
Test conditions are designed to provide different difficulties, with respect to such decision points. 
The test conditions are as follows:
\begin{enumerate}
	\item \emph{tmaze} (cf. Fig. \ref{fig:synth_data}-1): Observations prior to the junction yield no information about which side to choose. The predictor is expected to generate a prediction evenly distributed to both sides.
	\item \emph{tmaze-heavy-left} (Fig. \ref{fig:synth_data}-1): Similar to \emph{tmaze}, but the prediction to the left side should have higher weight (i.e. more particles should be located on the left side).
	\item \emph{tmaze-dirbias} (Fig. \ref{fig:synth_data}-2): Observations prior to the junction yield strong information about which side to choose, as the movement direction clearly indicates it. Given the LSTM-MDL has captured this property in the training phase, the predictor should generate paths which are clearly biased to one side.
	\item \emph{tmaze-posbias-gap} (Fig. \ref{fig:synth_data}-3): Similar to \emph{tmaze-dirbias} there is strong information about which side to choose in observations prior to the junction. Here, the absolute positioning of the observations indicate where to go. The gap between both directions should help the model to distinguish between the to classes.
	\item \emph{tmaze-posbias-nogap} (Fig. \ref{fig:synth_data}-4): Similar to \emph{tmaze-posbias-gap}, but there is no gap between both classes. That way, the model will start to mix predictions towards both sides when looking at observations located towards the center. With respect to the distribution of particles to either side, it should shift from one side to the other when looking at trajectories ranging from left to right.
\end{enumerate}
The biased test conditions resemble real world situations, where the decision on where to go at a junction can be indicated by the positioning and the movement direction.
Further, a t-shape was chosen as it provides a case where maximum likelihood predictors fail because it provides two possible outcomes, yet it is simple enough to ensure the reliability of the LSTM-MDL model.

\paragraph{Parameter configurations}
For the prediction algorithm, a fixed set of parameter configurations is evaluated.
Here, multinomial and stratified sampling strategies are tested.
Considering the weighting strategies, the following configurations are used:
\begin{enumerate}
	\item Unweighted (equal weights)
	\item Density value weighting
	\item Temperature weighting with $\tau \in \{0.01, 1000\}$
	\item Interpolation weighting with $\kappa \in \{0.25, 0.5, 0.75, 1\}$
\end{enumerate}
The parameters of temperature and interpolation weighting were chosen to cover a wide range of possible transformations with each strategy.
Combining the sampling and weighting strategies, yields a total of $16$ configurations.

\paragraph{Setup and Metric} 
In the following, the $N$-step prediction $\mathcal{P}^s$ (cf. algorithm \ref{alg:particle_prop}) given an observed trajectory fraction is evaluated.
For this, 50 trajectories are chosen from each synthetic test condition, such that their starting positions are evenly distributed from left to right.

For measuring the prediction quality, the distance between the particle distribution $\mathcal{P}^s$ and the expected distribution $D_{ex}$ is used.
$D_{ex}$ is obtained by selecting the endpoints of trajectories starting closely to the observed trajectory.
For $\mathcal{P}^s$, the first 15 positions are observed for each evaluation trajectory and the remaining $N$ positions are predicted (usually $N \in \left[50, 60\right]$).
As $\mathcal{P}^s$ and $D_{ex}$ are both sets of points, the distance is measured by calculating the euclidean distance between the corresponding centroids (average of all points):
\begin{align}
\text{CE} = ||\text{centroid}(\mathcal{P}^s) - \text{centroid}(D_{ex})||_2.
\end{align}

\begin{figure*}[t]
	\begin{center}		
		\includegraphics[width=0.3\textwidth]{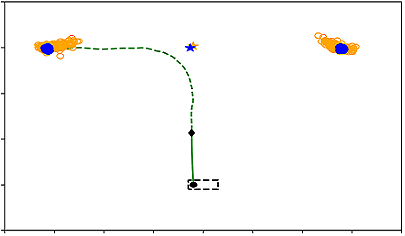}
		\includegraphics[width=0.3\textwidth]{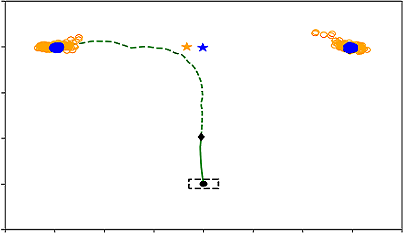}
		\includegraphics[width=0.3\textwidth]{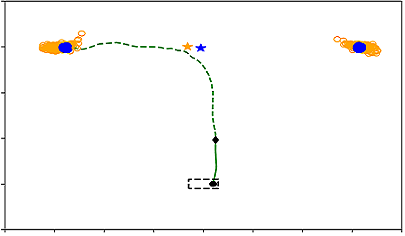}
		
		\includegraphics[width=0.3\textwidth]{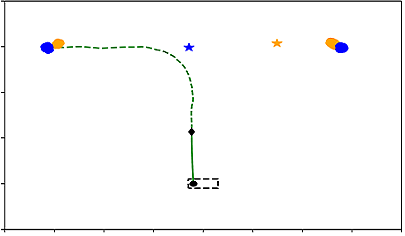}
		\includegraphics[width=0.3\textwidth]{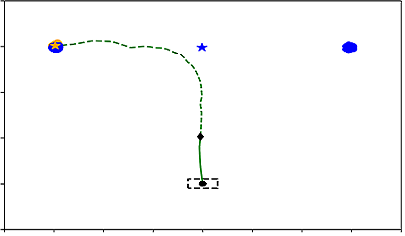}
		\includegraphics[width=0.3\textwidth]{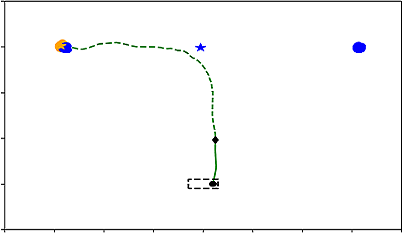}
	\end{center}
	\vspace{-0.2cm}
	\caption{Predictions for the endpoint distribution (orange) of different observed trajectories (solid green, dashed green is the true remainder), ground truth distributions (blue) and distribution centroids (stars). 
		First row: \emph{Unweighted multinomial} configuration.
		Second row: \emph{Density stratified} configuration.}
	\label{fig:centroid}
	\vspace{-0.1cm}
\end{figure*}
\paragraph{Comparison of configurations}
In a first step, the performance of different predictor configurations is compared.
Therefore, the centroid error for all test conditions and respective trajectories is averaged for each configuration (mean centroid error, MCE).
Additionally, the outlier ratio is calculated.
Here, the outlier ratio is the fraction of particles that are not within the bounds of all trajectory endpoints on the left or right side.
For calculating the MCE, outlier particles were disregarded.
The results are summarized in table \ref{tab:overall_res}.

\begin{table}
	\centering
	\begin{tabular}{l|l||c|c}
		sampling & weighting & MCE (std) & OR (std) \\  
		\hline
		\hline
		multinomial & - & \textbf{1.232} (2.484) & 0.018 (0.029) \\
		stratified & - & 1.242 (2.523) & 0.018 (0.029) \\
		multinomial & density  & 5.303 (6.135) & 0.000 (0.006) \\
		stratified & density  & 5.346 (6.195) & 0.000 (0.000) \\
		multinomial & $\tau = 0.01$ & 5.142 (7.295) & 0.073 (0.202) \\
		stratified & $\tau = 0.01$  & 5.241 (7.491) & 0.072 (0.205) \\
		multinomial & $\tau = 1000$ & 1.247 (2.522) & 0.017 (0.028) \\
		stratified & $\tau = 1000$ & 1.254 (2.543) & 0.017 (0.028) \\
		multinomial & $\kappa = 0.25$ & 1.239 (2.522) & 0.019 (0.030) \\
		stratified & $\kappa = 0.25$ & 1.242 (2.529) & 0.018 (0.029) \\
		multinomial & $\kappa = 0.5$ & 1.281 (2.589) & 0.019 (0.030) \\
		stratified & $\kappa = 0.5$ & 1.241 (2.524) & 0.018 (0.029) \\
		multinomial & $\kappa = 0.75$ & 1.294 (2.523) & 0.019 (0.030) \\
		stratified & $\kappa = 0.75$ & 1.242 (2.521) & 0.018 (0.029) \\
		multinomial & $\kappa = 1$ & 1.241 (2.556) & 0.018 (0.030) \\
		stratified & $\kappa = 1$ & 1.244 (2.522) & 0.019 (0.030) \\
		\hline
	\end{tabular}
	\caption{Mean centroid error (MCE) in meters, outlier ratio (OR) and standard deviations (std) for each configuration over all test conditions.
		Weighting: none (-), density value (density), temperature ($\tau =~?$) and interpolation ($\kappa =~?$).}
	\label{tab:overall_res}
	\vspace{-0.8cm}
\end{table}

It can be seen that not weighting the particles and using a multinomial sampling strategy performed best and density value weighting with stratified sampling performed worst.
Further, regardless of the sampling strategy, configurations using density value or temperature (with small $\tau$) weighting perform significantly worse than other configurations.
Also, the standard deviation of the results is much higher.
These configurations perform slightly better w.r.t. the outlier ratio.

\paragraph{Unbiased test conditions}
Following this, the cause for the discrepancy between configurations and the high standard deviation of worse configurations is examined.
Due to the similarities between configurations of either group, only \emph{unweighted multinomial} (best performance) and \emph{density stratified} (worst performance) are considered in the following.
While both configurations mostly work as expected on the biased test conditions, results differ on the unbiased test conditions.
Due to similar results for \emph{tmaze} and \emph{tmaze-heavy-left}, this evaluation is restricted to the \emph{tmaze} test condition.

Fig. \ref{fig:centroid} shows the prediction results for 3 trajectories (starting from the left, center and right of the starting region) for the \emph{unweighted multinomial} (top) and the \emph{density stratified} (bottom) configurations.
Here, the orange circles represent the prediction set $\mathcal{P}^s$, the blue circles the ground truth $D_{ex}$ and the stars the respective centroids. 
The \emph{unweighted multinomial} approach manages to predict both expected endpoint regions, but also causes the particles to spread more, which is the cause for a slightly higher outlier ratio.
A more important aspect visible in this figure, is that the particle set of the \emph{density stratified} configuration collapses in a dense set of particles, which often times leads to the prediction of only one of the two possible directions.
This particle set degeneration is the cause for higher centroid errors and an increased standard deviation.

\begin{figure}[h]
	\begin{center}		
		\includegraphics[width=0.45\textwidth]{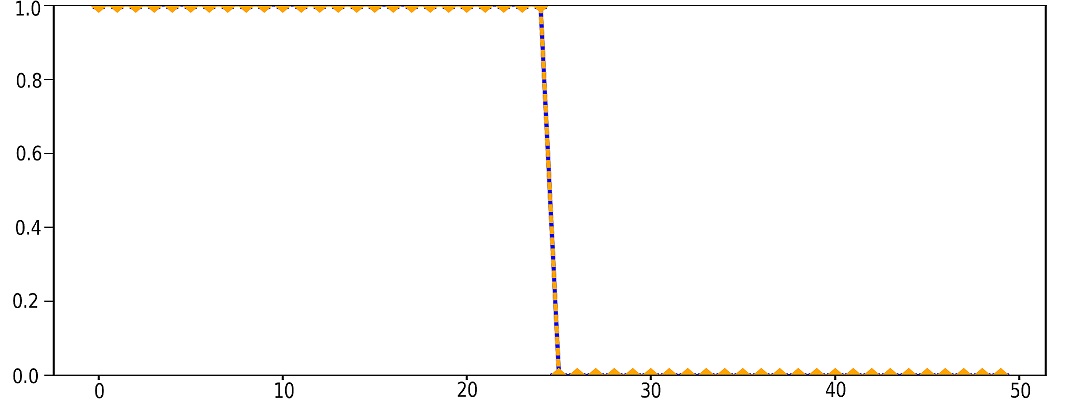}		
		\includegraphics[width=0.45\textwidth]{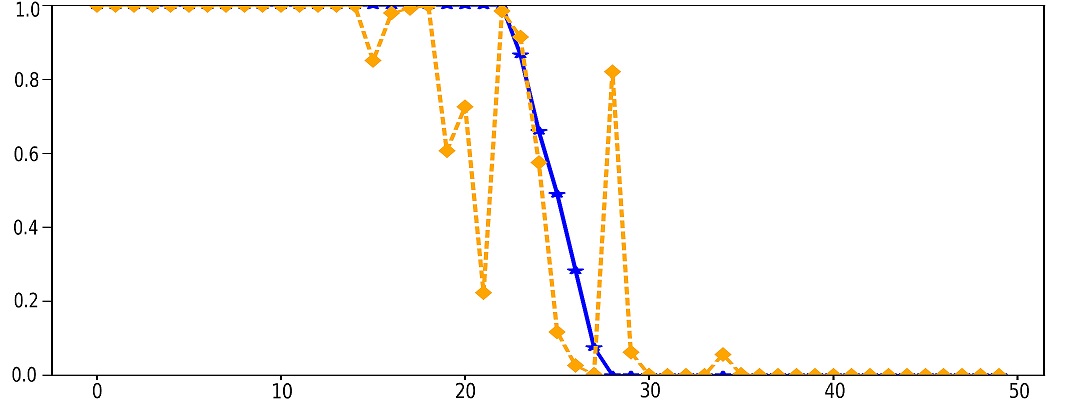}
	\end{center}
	\vspace{-0.2cm}
	\caption{Fraction of particles predicted to be in the left endpoint region for the \emph{unweighted multinomial} (yellow) configuration and all 50 evaluation trajectories of the \emph{tmaze-posbias-gap} test condition (top) and the \emph{tmaze-posbias-nogap} test condition (bottom). Blue plot: Ground truth.}
	\label{fig:prog}
	\vspace{-0.2cm}
\end{figure}

\paragraph{Distribution of particles}
Another interesting aspect exclusive to the \emph{tmaze(-heavy-left)} and \emph{gap} conditions is the ratio of particles located in the left and right region for the 50 selected trajectories.
This ratio can be expressed as the fraction of particles predicted to be inside the left endpoint region (disregarding outliers).
Considering \emph{tmaze} and \emph{tmaze-heavy-left}, this fraction is expected to be $0.5$ and $0.66$ on average for each of the 50 trajectories.
Here, the fractions generated by the \emph{unweighted multinomial} predictor are very close to the expected values, with average values of $0.54$ and $0.65$.
Fig. \ref{fig:prog} depicts the fraction of particles for each of the 50 trajectories selected from the gap test conditions.
For \emph{tmaze-posbias-gap}, a jump from a fraction of $1$ to $0$ is expected when passing the trajectories closest to the center of the starting region, as indicated by the ground truth (blue).                
As for \emph{tmaze-posbias-nogap}, a smooth transition from $1$ to $0$ is expected when passing the center-located trajectories.
In general, the \emph{unweighted multinomial} configuration manages to get close to these fractions (yellow).
Although the transition in \emph{tmaze-posbias-nogap} is not as smooth as desired, particles are still distributed to both sides for observations close to the center.

\paragraph{Conclusions}
The experiments show that configurations using density value or temperature (with small $\tau$) weighting suffer from particle set degeneration.
Although unsuitable for multi-modal path prediction, these configurations could be used for unimodal predictions.
These configurations, put strong emphasize on high weighted particles, thus a maximum likelihood prediction could be approximated.

As intended, given proper parameterization, the proposed sampling and weighting strategies enforce exploration of all paths known by the model.
Yet surprisingly, the simplest approach, calculating no weight and using a multinomial sampling approach, is on par with these configurations.
A possible explanation is that the LSTM-MDL model was reliable enough to properly capture motions seen in the training data.
Thus, the distribution output by $\mathcal{M}$ does not need to be manipulated artificially to put more emphasize on less likely modes.

\subsection{Qualitative results}
\label{ss:quali}

\begin{figure*}[t]
	\centering
	\begin{tabular}{l@{\hspace{3pt}}l@{\hspace{3pt}}r@{\hspace{3pt}}r}
		\includegraphics[width=0.24\textwidth]{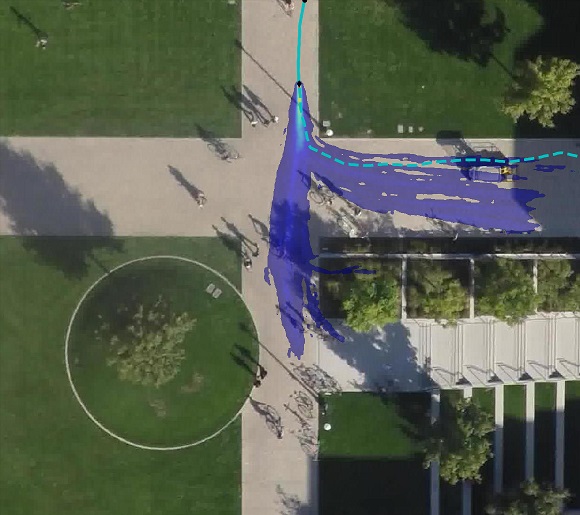} & \includegraphics[width=0.24\textwidth]{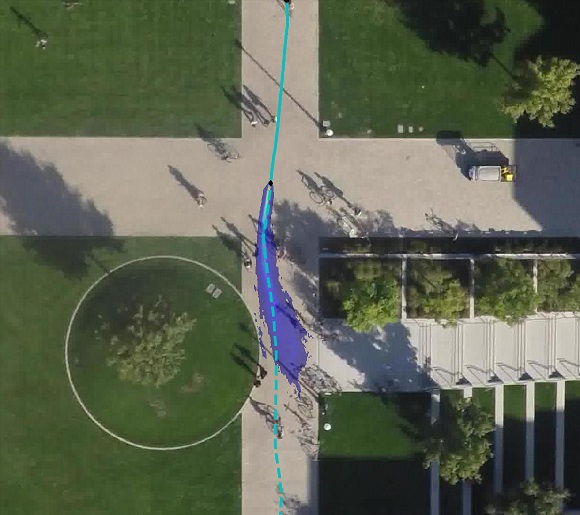} & \includegraphics[width=0.24\textwidth]{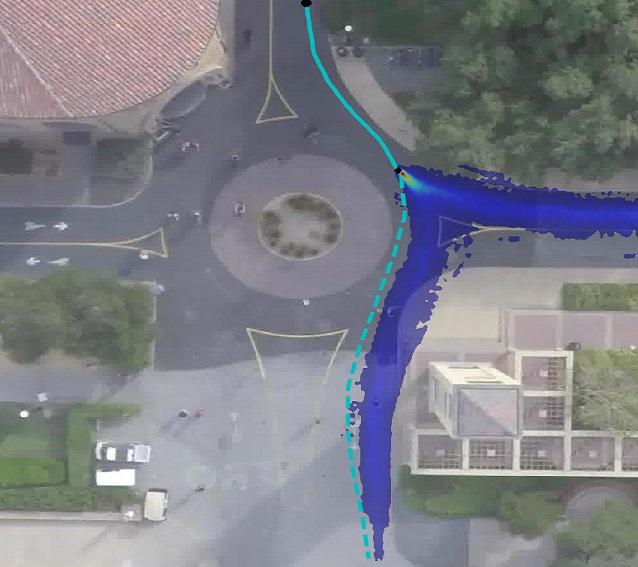} & \includegraphics[width=0.24\textwidth]{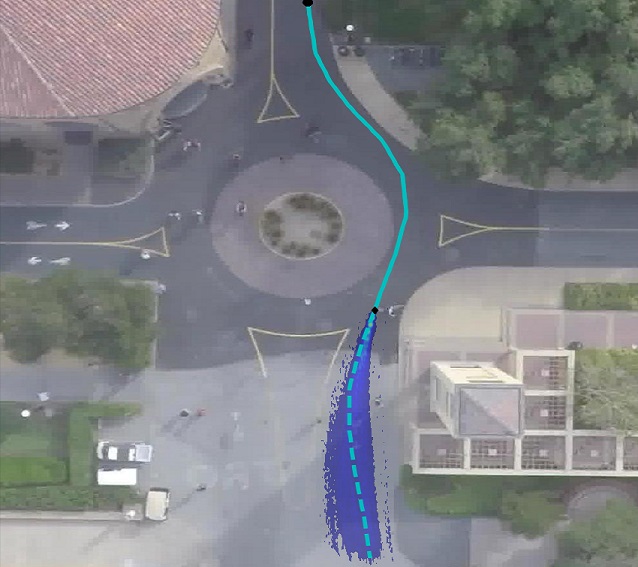} \\
		\small{~~1: 30 observed, 116 predicted} & \small{~~2: 60 observed, 102 predicted} & \small{3: 40 observed, 52 predicted~~~} & \small{4: 60 observed, 32 predicted~~~}
	\end{tabular}
	\caption{Prediction results (heatmap) for \emph{hyang} (1, 2) and \emph{deathcircle} (3, 4). 
			For each trajectory, a certain number of time steps is observed (solid cyan) and the remaining time steps (ground truth: dashed cyan) are predicted.
			The respective numbers are indicated in the subcaptions.} 
	\label{fig:qual_res}
	\vspace{-0.1cm}
\end{figure*}
In this section, the performance of the \emph{unweighted multinomial} configuration is shown on different real-world scenes.
For this, it is emphasized to use only datasets, that include at least one junction, thus leading to multiple hypotheses for future progressions of an observed trajectory.
Like this, scenarios are considered, where maximum likelihood predictors would fail because of their inability to generate multiple hypotheses.
This excludes popular datasets like \emph{ETH} \cite{pellegrini2009you} or \emph{UCY} \cite{lerner2007crowds}.
Instead, two scenes, taken from the \emph{Stanford Drone Dataset} \cite{robicquet2016learning}, are used: \emph{hyang} and \emph{deathcircle}.
In order to increase the amount of training data, the combined \emph{hyang} dataset only containing pedestrian trajectories, as provided by \cite{hug2017reliability}, was used.
Likewise, available video annotations for \emph{deathcircle} were combined into a single dataset by mapping image coordinates into a common reference frame using a homography projection calculated from 4 manually selected pixels, using the video's reference images.
It has to be noted, that although there are many pedestrians in the combined \emph{deathcircle} dataset, biker trajectories were used, as these follow the roads and the roundabout, leading to visually more inspectable results.
Further, as pedestrian interactions are rather short-termed events leading to small deviations from an intended path, these only have a limited influence on the trajectories on the regarded larger time scale.
Thus, this dataset can be used for evaluation, even with the LSTM-MDL model disregarding such interactions.

\begin{figure}[h]
	\begin{center}		
		\includegraphics[width=0.23\textwidth]{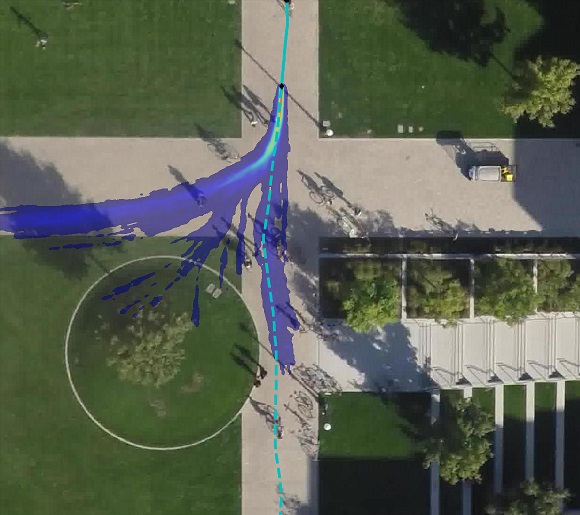}
		\includegraphics[width=0.23\textwidth]{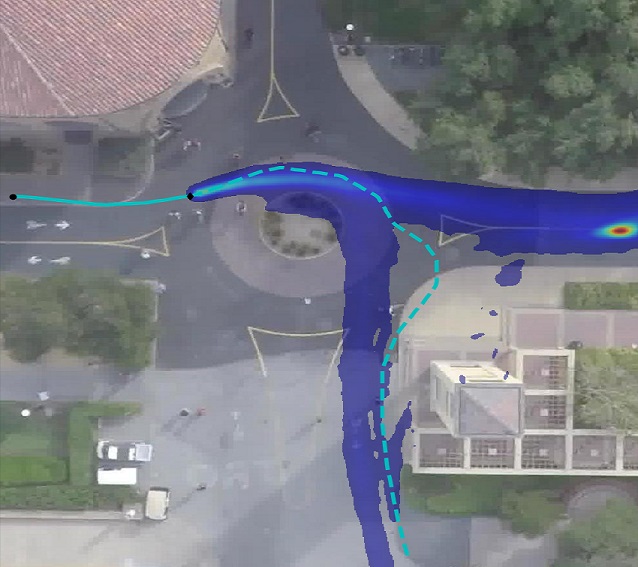}
	\end{center}
	\vspace{-0.2cm}
	\caption{Erroneous prediction results for \emph{hyang} (left; 30 steps observed, 132 predicted) and \emph{deathcircle} (right; 20 steps observed, 110 predicted).}
	\label{fig:qual_res2}
	\vspace{-0.25cm}
\end{figure}

\paragraph{Prediction results}
Exemplary prediction results are depicted in Fig. \ref{fig:qual_res} and \ref{fig:qual_res2}.
Here, the trajectory to be observed and predicted is illustrated in cyan (solid: observation, dashed: ground truth) and the prediction is illustrated as a heatmap calculated from the propagated particles.
Two exemplary trajectories are shown for \emph{hyang} (Fig. \ref{fig:qual_res}-1 and \ref{fig:qual_res}-2), where the first trajectory has been observed just before the junction and the second trajectory until it is extended into the junction.
Here, the predictor produces a prediction going straight over the junction and one heading towards the right side of the scene.
A prediction towards the left side is disregarded in this case due to the positioning of the observation close to the right edge of the pathway.
When the trajectory is observed for a longer period of time, the predictor disregards the prediction torwards the left or right side, and indicates that the trajectory is most likely to go downwards and eventually on the stairway.
Similar behavior can be observed for \emph{deathcircle}, where the predictor considers the correct paths to progress along.
In all cases, the ground truth lies within the predicted distribution.

\paragraph{Erroneous prediction results}
As stated in section \ref{s:path_pred}, the quality of the prediction also relies on the capabilities of the trained LSTM-MDL model.
Therefore, inaccuracies of the model are reproduced by the predictor, occasionally resulting in the generation of artifacts or missing predictions as depicted in Fig. \ref{fig:qual_res2}.
Looking at the prediction for a trajectory taken from \emph{hyang} (left) starting at the top, correct predictions (straight over the junction and towards the left side), but also a significant artifact (moving onto the grass) are produced.
As this artifact is located between the two correct paths, it most likely started as an outlier, hence entering a state, where less data was observed.
In such cases, the LSTM-MDL model presumably interpolates known motions, which results in a drift producing the artifact.
Besides this, when comparing this prediction to the prediction in Fig. \ref{fig:qual_res}-1, the previously indicated influence of the positioning of the observation on the prediction result is visible.
Here, the trajectory is located closer to the center of the pathway, making a prediction to the left side of the scene a viable option.
Due to the trajectory slightly heading towards the left side in the observation, a prediction to the right side is disregarded.

The right image in Fig. \ref{fig:qual_res2} shows a prediction containing artifacts for \emph{deathcircle}. 
Here, supposedly due to drifting, some particles are spread over the building in the bottom right portion of the scene.
Besides these artifacts, the predictor didn't identify the first exit of the roundabout as a possible progression for the trajectory.
This most likely stems from the fact, that only few trajectories in the training dataset actually take this route, making it statistically less relevant.
Also, in the last steps of the observation, the trajectory is very close to the center of the roundabout.
This may indicate that the trajectory is going straight or wrapping around the roundabout.

\section{Conclusions and outlook}
\label{s:conclusion}
In this paper, an approach to enable LSTM-MDL models to perform multi-modal pedestrian path prediction based on modified particle filter methods has been presented.
As this predictor can be stacked on top of any recurrent model with an MDL output layer, it could also be applied in different contexts.
In the experimental section, different predictor configurations have been tested using synthetic test conditions, in order to identify the best configurations to use.
These experiments show the counter-intuitive result of the simplest configuration being the best performing.
For this configuration, prediction results have been evaluated on different real-world scenes, concluding that the predictor is capable of producing plausible hypothesis on future paths.

Given the multi-modal prediction generated by the proposed approach, future works can elaborate on determining a global solution of a maximum likelihood prediction, which is different from a greedy solution in some cases.
Further, the prediction approach will be applied to more complex LSTM-MDL models which incorporate pedestrian interaction or local scene context.


\bibliographystyle{IEEEtran}
\bibliography{bibliography}

\begin{thebibliography}{10}
\providecommand{\url}[1]{#1}
\csname url@rmstyle\endcsname
\providecommand{\newblock}{\relax}
\providecommand{\bibinfo}[2]{#2}
\providecommand\BIBentrySTDinterwordspacing{\spaceskip=0pt\relax}
\providecommand\BIBentryALTinterwordstretchfactor{4}
\providecommand\BIBentryALTinterwordspacing{\spaceskip=\fontdimen2\font plus
\BIBentryALTinterwordstretchfactor\fontdimen3\font minus
  \fontdimen4\font\relax}
\providecommand\BIBforeignlanguage[2]{{%
\expandafter\ifx\csname l@#1\endcsname\relax
\typeout{** WARNING: IEEEtran.bst: No hyphenation pattern has been}%
\typeout{** loaded for the language `#1'. Using the pattern for}%
\typeout{** the default language instead.}%
\else
\language=\csname l@#1\endcsname
\fi
#2}}

\bibitem{huang2016deep}
S.~Huang, X.~Li, Z.~Zhang, Z.~He, F.~Wu, W.~Liu, J.~Tang, and Y.~Zhuang, ``Deep
  learning driven visual path prediction from a single image,'' \emph{IEEE
  Transactions on Image Processing}, vol.~25, no.~12, pp. 5892--5904, 2016.

\bibitem{bartoli2017context}
F.~Bartoli, G.~Lisanti, L.~Ballan, and A.~Del~Bimbo, ``Context-aware trajectory
  prediction,'' \emph{arXiv preprint arXiv:1705.02503}, 2017.

\bibitem{alahi2016social}
A.~Alahi, K.~Goel, V.~Ramanathan, A.~Robicquet, L.~Fei-Fei, and S.~Savarese,
  ``Social lstm: Human trajectory prediction in crowded spaces,'' in
  \emph{Proceedings of the IEEE Conference on Computer Vision and Pattern
  Recognition}, 2016, pp. 961--971.

\bibitem{kitani2012activity}
K.~M. Kitani, B.~D. Ziebart, J.~A. Bagnell, and M.~Hebert, ``Activity
  forecasting,'' in \emph{European Conference on Computer Vision}.\hskip 1em
  plus 0.5em minus 0.4em\relax Springer, 2012, pp. 201--214.

\bibitem{lee2017desire}
N.~Lee, W.~Choi, P.~Vernaza, C.~B. Choy, P.~H. Torr, and M.~Chandraker,
  ``Desire: Distant future prediction in dynamic scenes with interacting
  agents,'' 2017.

\bibitem{hochreiter1997long}
S.~Hochreiter and J.~Schmidhuber, ``Long short-term memory,'' \emph{Neural
  computation}, vol.~9, no.~8, pp. 1735--1780, 1997.

\bibitem{cho2014learning}
K.~Cho, B.~Van~Merri{\"e}nboer, C.~Gulcehre, D.~Bahdanau, F.~Bougares,
  H.~Schwenk, and Y.~Bengio, ``Learning phrase representations using rnn
  encoder-decoder for statistical machine translation,'' \emph{arXiv preprint
  arXiv:1406.1078}, 2014.

\bibitem{bishop2006pattern}
C.~M. Bishop, \emph{Pattern Recognition and Machine Learning (Information
  Science and Statistics)}.\hskip 1em plus 0.5em minus 0.4em\relax Secaucus,
  NJ, USA: Springer-Verlag New York, Inc., 2006.

\bibitem{graves2013generating}
A.~Graves, ``Generating sequences with recurrent neural networks,'' \emph{arXiv
  preprint arXiv:1308.0850}, 2013.

\bibitem{hug2017reliability}
R.~Hug, S.~Becker, W.~H{\"u}bner, and M.~Arens, ``On the reliability of
  lstm-mdl models for pedestrian trajectory prediction,'' \emph{VIIth
  International Workshop on Representation, analysis and recognition of shape
  and motion FroM Image data (RFMI 2017)}, 2017.

\bibitem{gu2015neural}
S.~Gu, Z.~Ghahramani, and R.~E. Turner, ``Neural adaptive sequential monte
  carlo,'' in \emph{Advances in Neural Information Processing Systems}, 2015,
  pp. 2629--2637.

\bibitem{zheng2017state}
X.~Zheng, M.~Zaheer, A.~Ahmed, Y.~Wang, E.~P. Xing, and A.~J. Smola, ``State
  space lstm models with particle mcmc inference,'' \emph{arXiv preprint
  arXiv:1711.11179}, 2017.

\bibitem{arulampalam2002tutorial}
M.~S. Arulampalam, S.~Maskell, N.~Gordon, and T.~Clapp, ``A tutorial on
  particle filters for online nonlinear/non-gaussian bayesian tracking,''
  \emph{IEEE Transactions on signal processing}, vol.~50, no.~2, pp. 174--188,
  2002.

\bibitem{kitagawa1996monte}
G.~Kitagawa, ``Monte carlo filter and smoother for non-gaussian nonlinear state
  space models,'' \emph{Journal of computational and graphical statistics},
  vol.~5, no.~1, pp. 1--25, 1996.

\bibitem{fishman2013monte}
G.~Fishman, \emph{Monte Carlo: concepts, algorithms, and applications}.\hskip
  1em plus 0.5em minus 0.4em\relax Springer Science \& Business Media, 2013.

\bibitem{hol2006resampling}
J.~D. Hol, T.~B. Schon, and F.~Gustafsson, ``On resampling algorithms for
  particle filters,'' in \emph{Nonlinear Statistical Signal Processing
  Workshop, 2006 IEEE}.\hskip 1em plus 0.5em minus 0.4em\relax IEEE, 2006, pp.
  79--82.

\bibitem{sutton1998reinforcement}
R.~S. Sutton and A.~G. Barto, \emph{Reinforcement learning: An
  introduction}.\hskip 1em plus 0.5em minus 0.4em\relax MIT press Cambridge,
  1998, vol.~1.

\bibitem{pellegrini2009you}
S.~Pellegrini, A.~Ess, K.~Schindler, and L.~Van~Gool, ``You'll never walk
  alone: Modeling social behavior for multi-target tracking,'' in
  \emph{Computer Vision, 2009 IEEE 12th International Conference on}.\hskip 1em
  plus 0.5em minus 0.4em\relax IEEE, 2009, pp. 261--268.

\bibitem{lerner2007crowds}
A.~Lerner, Y.~Chrysanthou, and D.~Lischinski, ``Crowds by example,'' in
  \emph{Computer Graphics Forum}, vol.~26, no.~3.\hskip 1em plus 0.5em minus
  0.4em\relax Wiley Online Library, 2007, pp. 655--664.

\bibitem{robicquet2016learning}
A.~Robicquet, A.~Sadeghian, A.~Alahi, and S.~Savarese, ``Learning social
  etiquette: Human trajectory understanding in crowded scenes,'' in
  \emph{European conference on computer vision}.\hskip 1em plus 0.5em minus
  0.4em\relax Springer, 2016, pp. 549--565.

\end{thebibliography}

\end{document}